\documentclass{Interspeech}

 \interspeechcameraready

\title{Improving Low-Resource Dialect Classification Using Retrieval-based Voice Conversion}

\author[affiliation={1}, equalcontribution]{Lea}{Fischbach}
\author[affiliation={2, 3}, equalcontribution]{Akbar}{Karimi}
\author[affiliation={1}]{Caroline}{Kleen}
\author[affiliation={1}]{Alfred}{Lameli}
\author[affiliation={2, 3}]{Lucie}{Flek}

\affiliation{Research Center Deutscher Sprachatlas}{Philipps-Universität Marburg}{Germany}
\affiliation{}{Lamarr Institute for ML and AI}{Germany}
\affiliation{}{b-it Center, University of Bonn}{Germany}

\email{lea.fischbach@uni-marburg.de, ak@bit.uni-bonn.de}


\usepackage{comment}

\usepackage[table]{xcolor}
\usepackage{array} 
\usepackage{makecell}
\usepackage{graphicx}
\usepackage{subcaption}
\usepackage{tabularx}

\definecolor{ageGreen}{RGB}{34,139,34}   
\definecolor{ageYellow}{RGB}{204,153,0}  
\definecolor{ageRed}{RGB}{180,0,0}       
\definecolor{ageViolet}{RGB}{138,43,226} 
\definecolor{goodGreen}{RGB}{46, 139, 87}
\definecolor{badRed}{RGB}{178, 34, 34}

\newcolumntype{s}{>{\hsize=.5\hsize}X}

\begin{document}

\maketitle

\begin{abstract}
\noindent
Deep learning models for dialect identification are often limited by the scarcity of dialectal data. To address this challenge, we propose to use Retrieval-based Voice Conversion (RVC) as an effective data augmentation method for a low-resource German dialect classification task. By converting audio samples to a uniform target speaker, RVC minimizes speaker-related variability, enabling models to focus on dialect-specific linguistic and phonetic features. Our experiments demonstrate that RVC enhances classification performance when utilized as a standalone augmentation method. Furthermore, combining RVC with other augmentation methods such as frequency masking and segment removal leads to additional performance gains, highlighting its potential for improving dialect classification in low-resource scenarios.

\end{abstract}

\section{Introduction}
\noindent
The scarcity of dialectal data poses a significant challenge for deep learning-based models in accurately identifying regional spoken dialects. Limited training data hinders the capabilities of these models to generalize across diverse speakers and dialectal variations. A common approach to address this issue is data augmentation, which increases both the quantity and diversity of training samples to improve the robustness of the model. Augmentation methods have proven effective in various speech-related tasks, including speech recognition \cite{Park_2019}, environmental sound classification \cite{madhu2019data}, keyword recognition \cite{wubet2022voice}, speaker verification \cite{du2021synaug} and speaker recognition \cite{yamamoto2019speaker, tao2024voiceconversionaugmentationspeaker}.

In this paper, we explore the effectiveness of state-of-the-art retrieval-based voice conversion (RVC) \cite{RVC} for dialect classification as a data augmentation method and investigate its potential as a complementary technique to traditional augmentation methods. Unlike conventional techniques that modify the audio signal without addressing speaker-related variability, RVC reduces such variability by converting all audio samples to a common target speaker, enabling the model to focus more effectively on linguistic and phonetic features specific to dialects. To verify this effect, we analyze how RVC modifies the audio signal, demonstrating that while speaker identity changes, critical prosodic features such as pitch and intonation remain stable. This is further supported by examining the resulting audio embeddings, which show reduced speaker-related clustering after conversion, confirming the role of RVC in minimizing speaker variability. Furthermore, we investigate whether matching the target speaker’s age to that of the source speakers influences classification performance, providing insights into the impact of speaker characteristics in RVC-based augmentation. Our results demonstrate that RVC not only improves model performance compared to standard methods like frequency masking but also leads to further gains when combined with them, offering a promising direction for dialect classification in low-resource settings.

\section{Related Work}
While traditional augmentation methods such as SpecAugment \cite{Park_2019} or SpecMix \cite{Kim2021SpecMixA} introduce variability at the spectral or temporal level, they do not address speaker-related variability, which can obscure dialect-specific features. This limitation is particularly relevant in dialect classification. Studies have shown that prosodic features, like pitch and intonation, are critical for dialect discrimination. Listeners can distinguish dialects using pitch cues, with classification accuracy improving when these cues are combined with rhythmic timing Information (Intonation) \cite{vicenik2013role}. Furthermore, comparative research indicates that German and English speakers exhibit narrower pitch spans and less pitch variability compared to Slavic languages like Bulgarian and Polish, suggesting that pitch range and intonation patterns are key markers of linguistic variation \cite{andreeva2014comparison}. 

Voice Conversion (VC) is commonly defined as modifying the speaker identity of a given utterance while preserving its linguistic content \cite{sisman2020overview}. 
Wubet and Lian \cite{wubet2022voice} demonstrated that VC-based augmentation outperforms baseline models that rely on affine transformations in speaker-independent keyword recognition. Shahnawazuddin et al. \cite{shahnawazuddin2020voice} reported significant reductions in word error rates (WERs) when augmenting children's speech data with VC. Singh et al. \cite{singh2021data} utilized a CycleGAN-based VC approach for end-to-end children’s ASR, showing that VC improves performance when combined with traditional augmentation methods like SpecAugment \cite{Park_2019} and speed perturbation, particularly in low-resource conditions. In the context of low-resource ASR, Baas and Kamper \cite{baas2022lowresource} found that VC consistently enhances recognition performance, especially when the available labeled data is extremely limited. However, as the amount of real data increases, the gains from VC plateau, suggesting diminishing returns in data-rich scenarios. For speaker recognition, Tao et al. \cite{tao2024voiceconversionaugmentationspeaker} proposed a nearest-neighbor VC augmentation strategy (VCA-NN), which improves performance in semi-supervised datasets by selecting source utterances close to the target speaker in a learned representation space. Meanwhile, Keskin et al. \cite{keskin2019measuringeffectivenessvoiceconversion} investigated how a CycleGAN-based VC model affects both speaker identification (SID) and ASR systems. They found that while VC models successfully mimic target speakers according to SID metrics, the improvements in ASR are marginal. Retrieval-based voice conversion (RVC) has recently been applied to low-resource ASR applications \cite{kamble2023custom, alhumud2024improving}, where the authors show that the quality of Hindi ASR \cite{kamble2023custom} and accented English ASR \cite{alhumud2024improving} is enhanced by RVC augmentation. The improvements by RVC in ASR pave the way for novel applications such as low-resource dialect classification to reap RVC's potential benefits.

\section{Method}
Using target speakers for the voice conversion model, we augment the original data to be used in our dialect classification pipeline. 

\subsection{Classification Pipeline}
\begin{figure}
    \centering
    \includegraphics[width=\columnwidth]{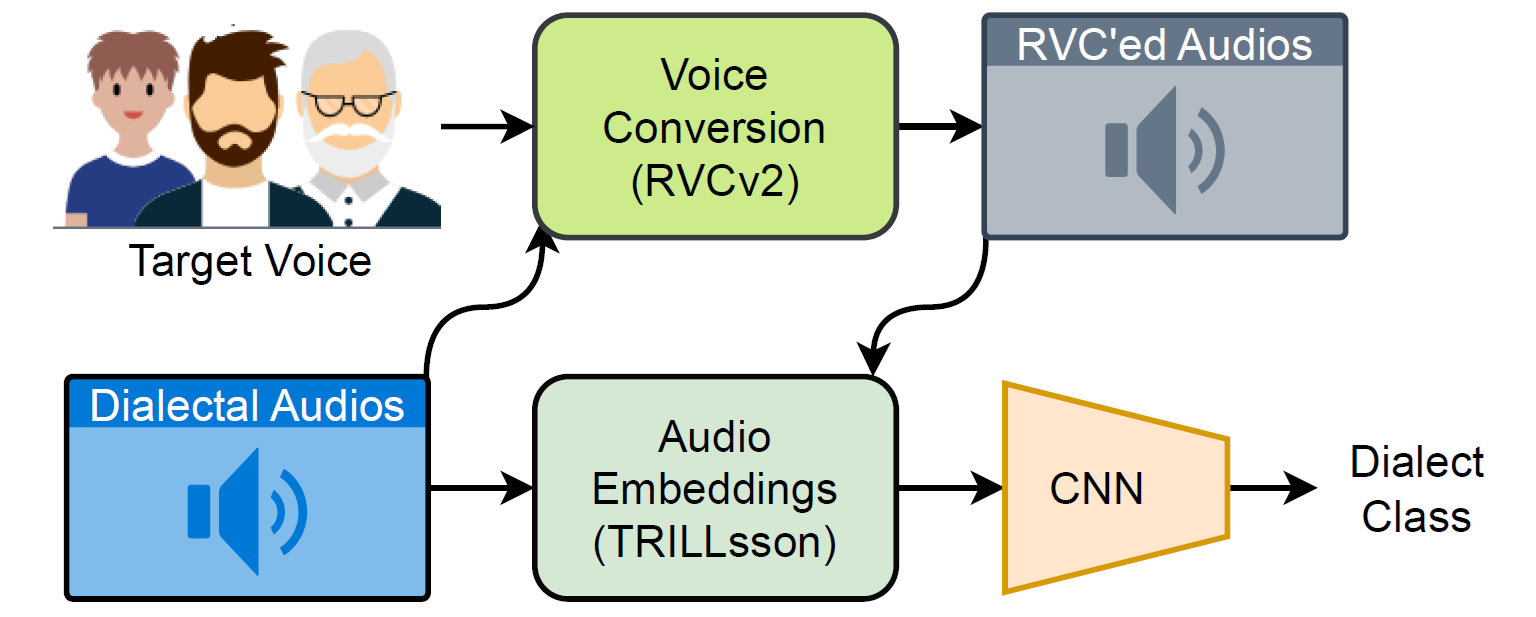}
    \caption{Dialect classification pipeline: We apply RVC to the original dialectal audios to generate RVCed samples. Both the original and converted audios are fed to the TRILLsson model to obtain audio embeddings, followed by a small CNN module.}
    \label{fig:pipeline}
\end{figure}

Our experimental pipeline\footnote{Detailed pipeline at \url{https://github.com/WoLFi22/DialectClassificationPipeline}.} processes original recordings by segmenting them into 10-second audio segments. These segments serve as inputs for Google’s TRILLsson models \cite{shor22_interspeech}, which extract high-level embedding vectors. To ensure the model does not memorize specific speakers, we employ a train-validation-test split such that each speaker appears in exactly one partition. For the validation and test sets, we randomly select $\lceil\frac{\#S_D}{10}\rceil$ speakers of each dialect, where $\#S_D$ represents the total number of speakers in the respective dialect. Since speaker selection can directly influence overall results, we conduct 250 runs using different random subsets. This large number of runs provides statistical robustness and yields a more reliable estimate of real-world performance. Any significance testing between pairs of runs is carried out via the Mann-Whitney U test \cite{Mann1947}. The CNN module follows a straightforward feedforward design with two hidden layers, each followed by a LeakyReLU activation and a dropout layer. The output layer uses a softmax function for classification. Figure \ref{fig:pipeline} illustrates at a high-level the pipeline used for experiments involving voice-converted audio samples, which extends the standard processing pipeline described above. We report the mean weighted F1 score as our primary evaluation metric.

\subsection{Data Augmentation Techniques}
\subsubsection{Retrieval-based Voice Conversion (RVC)}
\noindent
For voice conversion, we utilized RVCv2\footnote{\url{https://github.com/RVC-Project/Retrieval-based-Voice-Conversion-WebUI}} model which is widely used for retrieval-based voice conversion. Its main component is the VITS model \cite{kim2021conditional} which is an end-to-end text-to-speech (TTS) framework. Using normalizing flows \cite{rezende2015variational} and an adversarial training process \cite{goodfellow2014generative}, the VITS model enables RVCv2 to produce natural-sounding audio.

\subsubsection{Traditional Baseline Augmentation}
\noindent
Along with voice conversion as an augmentation method, we also consider the impact of two other widely used augmentation methods, namely Frequency Masking (FM) and Segment Removal (SR). In our augmentation pipeline, SR is applied first, followed by FM. For SR, we apply 50\% augmentation with 0.3-second augmentation intervals for each audio segment, resulting in 16 removed chunks (4.8 seconds total) per 10-second audio segment. This approach helps reduce the number of training samples, thus decreasing computational costs and training time while maintaining overall model performance. For FM, we randomly generate between 1 and 3 non-overlapping masking intervals per audio segment, with each interval targeting a frequency band selected randomly within a bandwidth range of 100 to 2500 Hz. These masking operations span the entire audio segment, introducing variability by affecting different spectral regions.

\section{Experiments and Results}
We performed our experiments using TensorFlow as the primary framework, running on an NVIDIA GeForce RTX 3080 Ti laptop GPU equipped with 16 GB of dedicated memory. The system was configured on an Intel 64-bit Windows 11 operating system, with the CUDA Toolkit and NVIDIA cuDNN libraries installed to facilitate GPU-accelerated deep learning applications.

\subsection{Dataset}
Spoken dialect datasets are rare and most of the machine learning research has focused on text-based dialectal data \cite{joshi2024natural}. One of the available spoken dialect datasets is the "Regionalsprache.de" (REDE) corpus \cite{rede}, which includes recordings from 5 speaking situations, 593 speakers, and 115 locations across Germany. Its broad regional and speaker diversity makes it a unique resource for dialect research. For this study, we selected the translation of the "Wenker Sentences" \cite{Wenker-Phrases} into local dialects. In this task, speakers translated 40 standard German sentences into their local dialects, providing consistent and comparable linguistic material across different regions\footnote{Additional information about the recording conditions, locations and the REDE project can be found at \url{https://rede-infothek.dsa.info/}.}. The dataset includes speakers from three age groups: \textit{Young} (18–23 years), \textit{Middle}-aged (45–55 years), and \textit{Old} (65 years and older).
The raw dataset contains 164.62 hours of audio, including pauses and speech from non-target speakers. To prepare the data for our classification task, the audio was first preprocessed, which involved downsampling to 16 kHz, with 16-bit resolution in mono format. After preprocessing, we applied speaker diarization to identify and separate individual speakers within the audio recordings. Based on the comparative analysis of different diarization tools presented in \cite{fischbach2024comparative}, we selected the NeMo toolkit \cite{kuchaiev2019nemo} due to its demonstrated performance in that study. To ensure data quality, we excluded segments shorter than one continuous second after the diarization process.
After applying speaker diarization, we obtained 43.71 hours of dialectal speech, which serves as the final dataset for our experiments.

Details regarding the distribution of speakers across age groups, the number of samples, and the number of speakers for validation and testing can be found in Table \ref{tab:dialect-overview-all}. In addition to individual age groups, we consider the category of \textit{All}, which refers to the combination of speakers from all three age groups. This allows us to analyze both age-group-specific performance and overall trends throughout the dataset.
The speakers are classified into 20 distinct dialect groups according to Wiesinger \cite{wiesinger1983einteilung}. Transitional areas between dialects were not considered to maintain clear boundaries between categories. Moreover, only dialects with at least three available speakers were included in the analysis to ensure sufficient data for reliable classification.
\begin{table}[t]
\caption{Overview of dataset distribution across the age groups.}
\centering
\footnotesize
\setlength{\tabcolsep}{3pt}
\rowcolors{2}{gray!15}{white}
\begin{tabularx}{\columnwidth}{X c r c c}
\toprule
Age group & \makecell[l]{\# Speakers\\(Total)} & Total Seconds & \# Samples & \makecell[l]{\# Speakers\\(Val/Test)}\\
\midrule
 \textcolor{ageGreen}{Young}
    & 139 & 32440.17  & 3170 & 22 \\
 \textcolor{ageYellow}{Middle}
    & 237 & 59966.90  & 5875 & 32 \\
 \textcolor{ageRed}{Old}
    & 198 & 64476.90  & 6352 & 29 \\
 \textcolor{ageViolet}{All}
    & 574 & 156883.97 & 15397 & 69 \\
\end{tabularx}
\label{tab:dialect-overview-all}
\end{table}

\subsection{Experimental Setup}
We compare four experimental conditions: the baseline, which uses only the original audio samples; SR-FM augmentation, where a combination of Segment Removal (SR) and Frequency Masking (FM) is applied; RVC augmentation, which incorporates voice-converted samples as additional training data; and a combination of RVC + SR-FM augmentation, where both augmentation methods are used together. For SR-FM augmentation, we evaluated two settings: generating one augmented file per original sample (SR-FM-1) and six augmented files per original sample (SR-FM-6). The choice of six augmented files was based on preliminary experiments, which showed that increasing the number of augmented samples enhances model performance due to greater data diversity. Importantly, original samples are always included in the training set and used for evaluation/testing, regardless of the applied augmentation strategy. 
In addition, we conducted experiments separately for each age group of speakers as well as for all age groups combined. For RVC, we initially used a middle-aged target speaker for all recordings (which we refer to as RVC-1). To further investigate the effect of target speaker characteristics, we also conducted experiments using one target speaker per age group (RVC-3): the same middle-aged target speaker as for RVC-1 for the middle-aged group, a younger target speaker for the younger age group, and an older target speaker for the elderly group, each selected to match the corresponding age group.

\subsection{Results and Analysis}
\subsubsection{Performance Evaluation}
\begin{table*}[ht]
    \caption{Comparison of classification results for different augmentation strategies and age groups. The table lists the tested condition, the original and augmented sample counts, the reference condition, the p-value of the Mann–Whitney U test, and the F1 score.}
    \centering
    \footnotesize
    \rowcolors{2}{gray!15}{white}
    \begin{tabular}{llllcclllccc}
        \toprule
        Row & \multicolumn{3}{c}{Test} & \makecell[l]{\# Original\\Samples} & \makecell[l]{\# Augmented\\Samples} & \multicolumn{3}{c}{Tested against} & \makecell[l]{p-value\\U-Test} & \multicolumn{2}{c}{\makecell[l]{Mean\\weighted F1}} \\
        \midrule
        1 & & Baseline & \textcolor{ageGreen}{Young} & 3170 & - & & - &  & - & 0.148 & $\pm$ 0.040 \\
        2 & RVC-1 & & \textcolor{ageGreen}{Young} & 3170 & 3170 & & SR-FM-6 & \textcolor{ageGreen}{Young} & \textcolor{goodGreen}{0.000} & 0.180 & $\pm$ 0.047 \\
        3 & RVC-1 & SR-FM-6 & \textcolor{ageGreen}{Young} & 3170 & 9516 & RVC-1 & & \textcolor{ageGreen}{Young} & \textcolor{goodGreen}{0.004} & 0.193 & $\pm$ 0.049 \\
        4 & RVC-3 & SR-FM-6 & \textcolor{ageGreen}{Young} & 3170 & 9516 & RVC-1 & SR-FM-6 & \textcolor{ageGreen}{Young} &  0.618 & 0.195 & $\pm$ 0.044 \\
        5 & & Baseline & \textcolor{ageYellow}{Middle} & 5875 & - & & - &  & - & 0.348 & $\pm$ 0.052 \\
        6 & RVC-1 & & \textcolor{ageYellow}{Middle} & 5875 & 5875 & & SR-FM-6 & \textcolor{ageYellow}{Middle} & \textcolor{goodGreen}{0.000} & 0.376 & $\pm$ 0.048 \\
        7 & RVC-1 & SR-FM-1 & \textcolor{ageYellow}{Middle} & 5875 & 2928	& RVC-1 & & \textcolor{ageYellow}{Middle} & 0.599 & 0.379 & $\pm$ 0,047 \\
        8 & RVC-1 & SR-FM-6 & \textcolor{ageYellow}{Middle} & 5875 & 17568 & RVC-1 & & \textcolor{ageYellow}{Middle} & \textcolor{goodGreen}{0.001} & 0.391 & $\pm$ 0.047 \\
        9 & & Baseline & \textcolor{ageRed}{Old} & 6352 & - & & - &  & - & 0.321 & $\pm$ 0.057 \\
        10 & RVC-1 & & \textcolor{ageRed}{Old} & 6352 & 6352 & & SR-FM-6 & \textcolor{ageRed}{Old} & \textcolor{goodGreen}{0.000} & 0.351 & $\pm$ 0.055 \\
        11 & RVC-1 & SR-FM-6 & \textcolor{ageRed}{Old} & 6352 & 19188 & RVC-1 & & \textcolor{ageRed}{Old} & \textcolor{goodGreen}{0.013} & 0.362 & $\pm$ 0.055 \\
        12 & RVC-3 & SR-FM-6 & \textcolor{ageRed}{Old} & 6352 & 19188 & RVC-1 & SR-FM-6 & \textcolor{ageRed}{Old} & 0.728 & 0.365 & $\pm$ 0.053 \\
        13 & & Baseline & \textcolor{ageViolet}{All} & 15397 & - & & - &  & - & 0.404 & $\pm$ 0.040 \\
        14 & RVC-1 & & \textcolor{ageViolet}{All} & 15397 & 15397 & & SR-FM-6 & \textcolor{ageViolet}{All} & 0.379 & 0.412 & $\pm$ 0.044 \\
        15 & RVC-1 & SR-FM-1 & \textcolor{ageViolet}{All} & 15397 & 7712 & RVC-1 & & \textcolor{ageViolet}{All} & \textcolor{goodGreen}{0.006} & 0.422 & $\pm$ 0.041 \\
        16 & RVC-1 & SR-FM-6 & \textcolor{ageViolet}{All} & 15397 & 46272 & RVC-1 & & \textcolor{ageViolet}{All} & \textcolor{goodGreen}{0.013} & 0.420 & $\pm$ 0.036 \\
        17 & RVC-3 & SR-FM-6 & \textcolor{ageViolet}{All} & 15397 & 46272 & RVC-1 & SR-FM-6 & \textcolor{ageViolet}{All} & 0.212 & 0.425 & $\pm$ 0.040 \\
        \bottomrule
    \end{tabular}
    \label{tab:results}
\end{table*}

\begin{figure}
    \centering
    \includegraphics[width=\columnwidth]{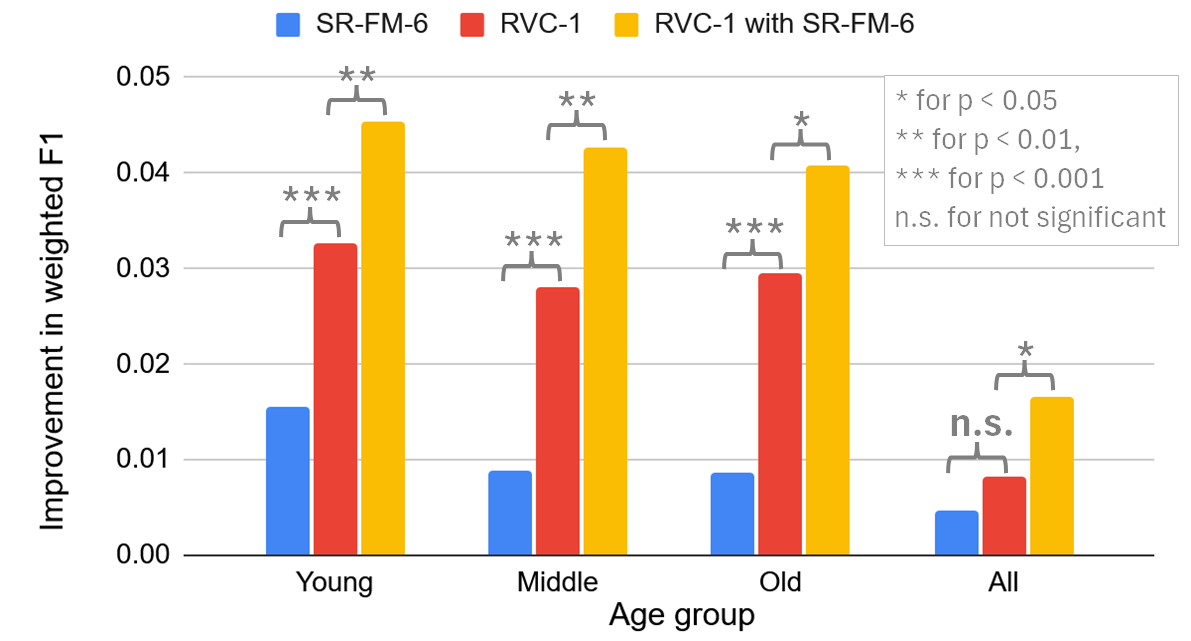}
    \caption{In addition to being advantageous when used alone, RVC further improves performance when combined with segment removal (SR) and frequency masking (FM) augmentations.}
    \label{fig:overall-improvement}
\end{figure}

\noindent
Figure \ref{fig:overall-improvement} illustrates the performance gains from applying voice conversion (RVC-1) for dialect classification. The statistical significance of the improvements is indicated directly in the figure using asterisk markers (e.g., * for $p < 0.05$). All improvements shown are absolute differences in weighted F1-score relative to the baseline (original samples only), with mean weighted F1-scores of 0.148, 0.348, 0.321 for individual generations and 0.404 for all generations combined. For comparison, we also include results from traditional augmentation techniques, specifically Segment Removal (SR) and Frequency Masking (FM), combined to produce six augmented files per original sample (SR-FM-6). Overall, we observe absolute improvements of up to 0.03 in weighted F1-score for individual generations when using RVC-1, and up to 0.045 when combined with SR-FM-6. However, the impact is less pronounced considering all generations combined, likely due to the increased speaker variability, which reduces the relative gains from augmentation. 

Detailed results for our experiments, including the exact p-values for all comparisons, are given in Table \ref{tab:results}. Comparing RVC with other augmentations, RVC-1 performs significantly better than SR-FM-6 across all individual age groups (Rows 2,6,10), although this difference is not significant when considering all age groups combined (Row 14). While SR-FM-6 requires multiple augmented files, RVC-1 achieves comparable or better performance with fewer samples, offering more efficient model training. However, this comes with the trade-off of additional computational overhead during the generation of voice-converted audio files. Combining RVC-1 with SR-FM-6 yields further improvements, significantly outperforming RVC-1 alone (Rows 3,8,11,16). This highlights the complementary nature of the RVC augmentation technique. 

For a fair comparison, we also evaluated SR-FM with only one augmented file per original sample (SR-FM-1). Unlike SR-FM-6, there is no significant performance difference between RVC-1 alone and RVC-1 + SR-FM-1 for the middle-aged (Row 7) and older groups, while performance is significantly worse for the younger one. For all age groups combined, the combination is still significantly better (Row 15), suggesting that the benefits of combining SR-FM with RVC are more pronounced when multiple augmented files are generated. Additionally, to investigate the impact of target speaker characteristics, we conducted experiments where we assigned age-matched target speakers to the source speakers, marked as RVC-3 in Table \ref{tab:results}. The aim was to assess whether matching the target speaker’s age to that of the source speaker influences classification performance. The results show no significant performance difference compared to using a single middle-aged target speaker, both for individual age groups and for the combined dataset (Rows 4,12,17). This indicates that the age of the target speaker does not substantially affect the effectiveness of RVC for dialect classification.

\subsubsection{How RVC Changes the Voice}
\begin{figure}
    \centering
    \includegraphics[width=\linewidth]{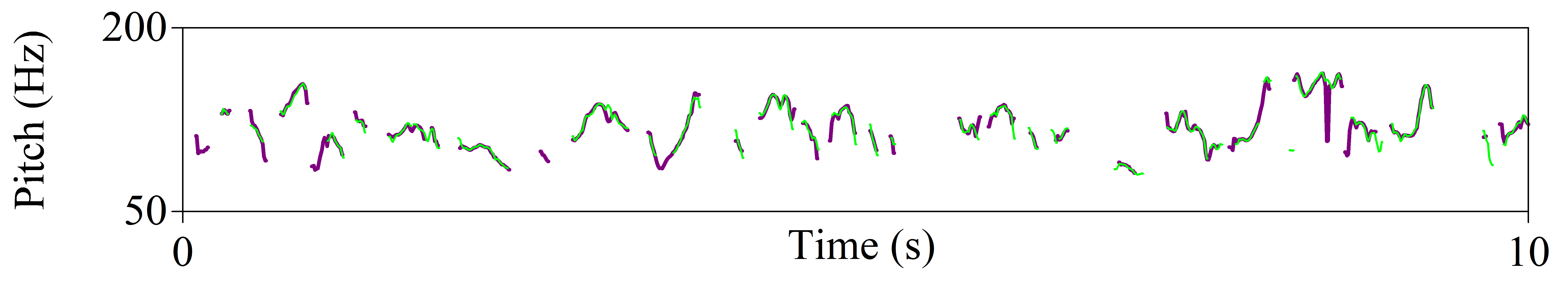}
    \caption{Modification of pitch by retrieval-based voice conversion (RVC). Purple is the pitch of the original sample and green the pitch of the same sample when RVC is applied.}
    \label{fig:pitch}
\end{figure}

\noindent
To understand how retrieval-based voice conversion (RVC) modifies speech, we analyzed key acoustic features, focusing on the pitch (F0), the first three formant frequencies (F1, F2, F3) and the resulting audio embeddings. As shown in Figure \ref{fig:pitch}, the pitch contours of original (purple) and voice-converted (green) samples show minimal differences, indicating that RVC does not significantly affect intonation, temporal structure, or rhythmic patterns. Pitch extraction was performed using Praat \cite{praat}. A statistical analysis across all samples confirmed that the mean pitch remains virtually unchanged (118.73 $\pm$ 10.42 Hz to 118.29 $\pm$ 10.38 Hz).

While pitch remains stable, RVC introduces changes in formant structure, which are key to shaping speaker identity. Comparative analysis shows that mean F1 increased from 535.71 Hz to 624.39 Hz, F2 decreased slightly from 1626.20 Hz to 1598.07 Hz and F3 shifted from 2616.61 Hz to 2666.23 Hz. Additionally, the variability of these formants, especially F3, decreased significantly (from 92.26 to 40.99), indicating more standardized acoustic patterns. While the overall formant structure is preserved, these shifts likely contribute to perceptual changes in timbre, altering the perceived voice without distorting core linguistic content. This transformation effectively changes vocal identity while maintaining the acoustic features critical for conveying dialectal information.

\begin{figure}
    \centering
    \includegraphics[width=\linewidth]{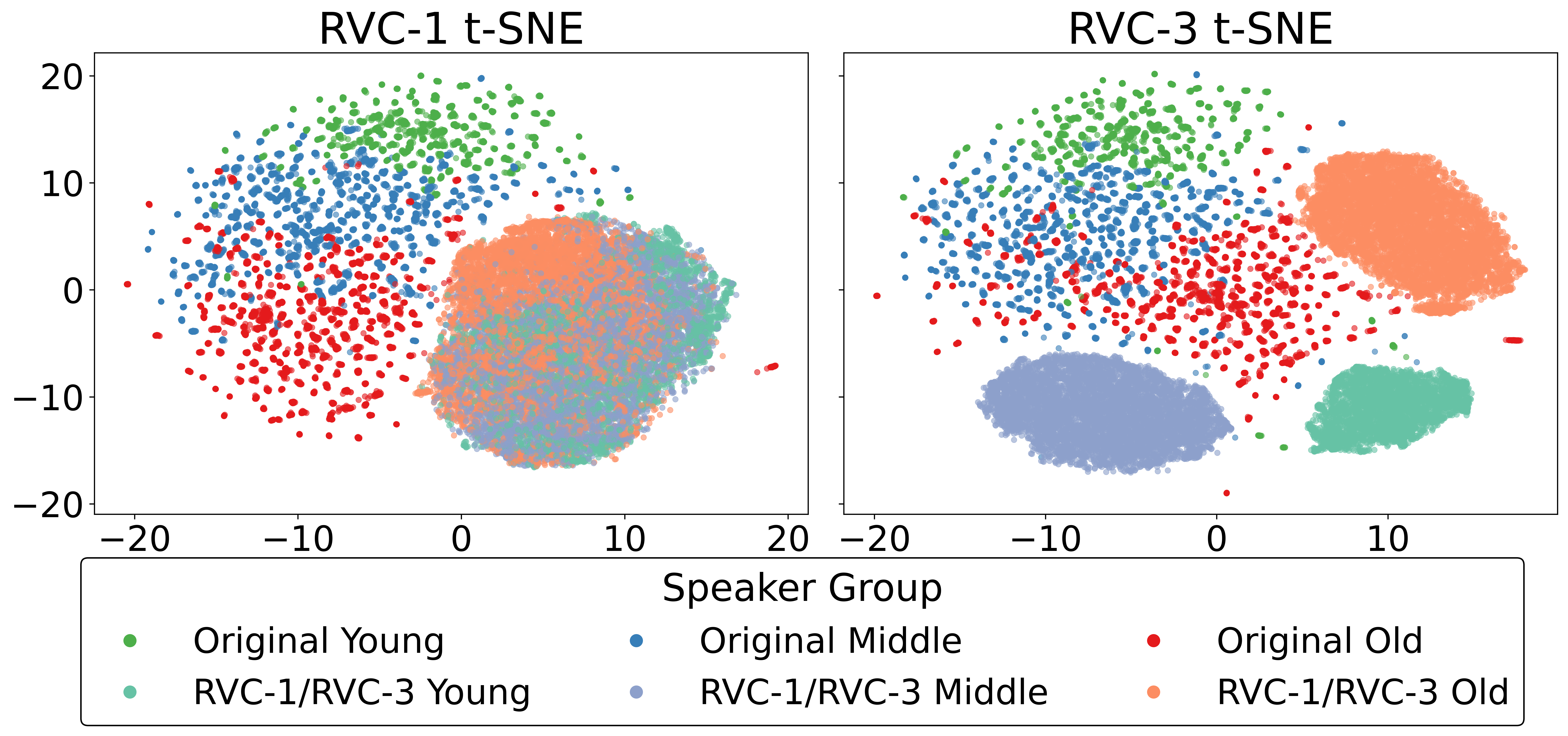}
    \caption{Original and voice-converted embeddings (best seen in color). RVC-1 refers to conversion to a single target speaker, while RVC-3 involves conversion to different target speakers for each age group matching the respective age group.}
    \label{fig:embeddings}
\end{figure}

The impact of RVC on speaker characteristics is further illustrated in the t-SNE plots in Figure \ref{fig:embeddings}, which visualize the distribution of audio embeddings before and after conversion. In the left plot, original samples and their RVC-1 counterparts (converted to the same target speaker) show that original samples form distinct clusters based on both individual speakers and age groups. However, after applying RVC, these clusters merge, indicating that the embeddings are no longer dominated by speaker identity. The plot on the right compares the original samples with those processed through RVC-3, where each age group is converted to a different target speaker. In this case, we observe the formation of three distinct clusters for RVC-3, corresponding to each of the generation-specific target speakers, but still without any remaining speaker-specific sub-clusters.

\section{Conclusion}
\noindent
We investigated the use of retrieval-based voice conversion (RVC) as a data augmentation technique to improve low-resource dialect classification. Our results demonstrate that RVC is an effective augmentation method and can also be combined with simple augmentation techniques such as frequency masking and segment removal, leading to additional performance improvements. This highlights the complementary nature of these techniques, where RVC addresses speaker variability while traditional methods introduce data diversity. One key observation is that RVC effectively alters vocal identity without distorting the core acoustic features that convey dialectal information. Additionally, we explored the impact of using multiple target speakers that match age groups, with findings that indicate no significant difference in the performance between single and multiple target speakers for RVC.

\section{Acknowledgements}
This research is supported by the Academy of Science and Literature Mainz (grant REDE 0404), the German Federal Ministry of Education and Research (BMBF) (grant AnDy 16DKWN007), the state of North-Rhine Westphalia as part of the Lamarr-Institute for Machine Learning and Artificial Intelligence, LAMARR22B and the Research Center Deutscher Sprachatlas Marburg. \\

\bibliographystyle{IEEEtran}
\bibliography{mybib}

\end{document}